\documentclass[10pt,twocolumn,letterpaper]{article}

\usepackage{wacv}

\usepackage{graphicx}
\usepackage{booktabs}
\usepackage{amsmath,amssymb}
\usepackage{xcolor}
\usepackage{xspace}

\newcommand{\method}{DensePol\xspace}
\newcommand{\dg}{\ensuremath{^{\circ}}}

\AtBeginDocument{
\setlength{\abovedisplayskip}{4pt}
\setlength{\belowdisplayskip}{4pt}
}

\definecolor{wacvblue}{rgb}{0.21,0.49,0.74}
\usepackage[pagebackref,breaklinks,colorlinks,allcolors=wacvblue]{hyperref}

\def\wacvPaperID{784} 
\def\confName{WACV}
\def\confYear{2027}

\title{DensePol: Dense-Angle Polarization Dataset for Learning-Based Polarimetric Vision}

\author{
Param Sangani\textsuperscript{1} \quad
Ahmad Moori\textsuperscript{1} \quad
Erik Blasch\textsuperscript{2} \quad
Guna Seetharaman\textsuperscript{3} \quad
Hadi AliAkbarpour\textsuperscript{1}\\
\textsuperscript{1}Department of Computer Science,
Saint Louis University, St.~Louis, MO, USA\\
\textsuperscript{2}MOVEJ Analytics, Fairborn, OH, USA\\
\textsuperscript{3}U.S.~Naval Research Laboratory,
Washington, DC, USA\\
{\small\ttfamily
\{param.sangani,ahmad.moori,hadi.akbarpour\}@slu.edu}\\
{\small\ttfamily
erik.blasch.civ@us.af.mil \quad
guna.seetharaman@nrl.navy.mil}
}

\begin{document}
\maketitle

\begin{abstract}

Polarimetric vision is gaining increasing attention because it provides physical cues about scene shape, material, and reflection that are difficult to recover from RGB alone. Recent work has therefore explored predicting polarization directly from conventional RGB images; however, the fidelity of these methods strongly depends on the polarization supervision used for training. Most existing datasets rely on Division-of-Focal-Plane (DoFP) cameras with four spatially interleaved analyzer orientations, which provide limited angular redundancy and introduce interpolation and instantaneous-field-of-view errors. We introduce DensePol, a high-redundancy RGB--polarization dataset based on Division-of-Time (DoT) acquisition, capturing 180 full-resolution analyzer orientations at $1^\circ$ intervals. DensePol contains 2,018 paired RGB--polarization images with the angular measurements and fitting residuals retained. Dense angular sampling substantially improves polarization stability, reducing AoLP deviation from $13.36^\circ$ to $2.21^\circ$. We further introduce a deterministic diffusion-based RGB-to-polarization framework with cyclic AoLP representation and a local DoLP refiner. Experiments demonstrate improved polarization prediction and downstream surface-normal estimation. The dataset and code will be publicly available.

\end{abstract}

\section{Introduction}
\label{sec:intro}
\begin{figure}[t]
  \centering
  \includegraphics[width=\linewidth]{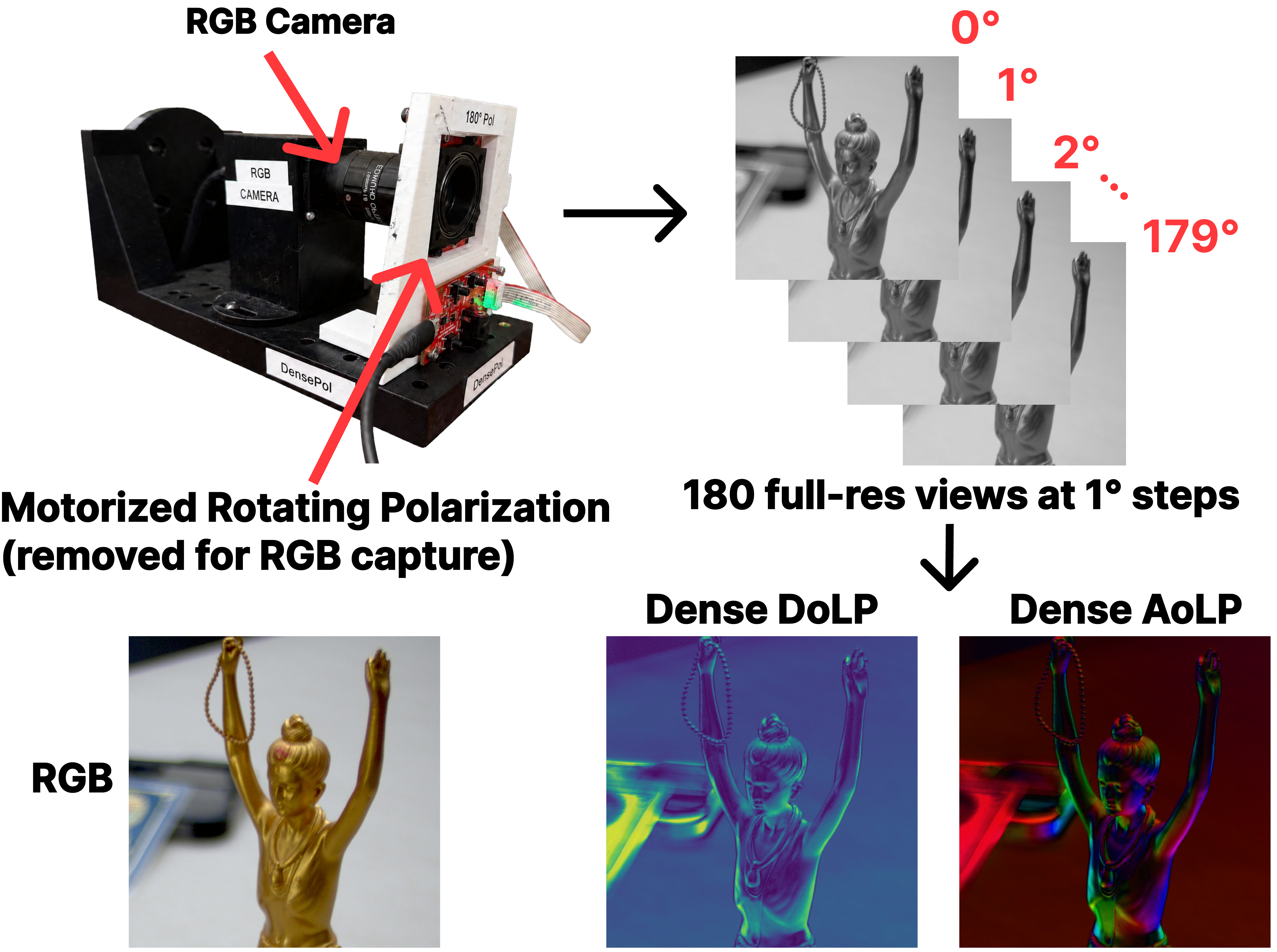}
  \caption{\textbf{Overview of DensePol.} A motorized polarizer captures 180 full-resolution views at $1\dg$ intervals to generate dense DoLP and AoLP references, together with a separate polarizer-free RGB image.}
  \label{fig:overview}
  \vspace{-14pt}
\end{figure}
Polarization exposes surface, reflection, and material cues that are weak or absent in RGB, supporting shape estimation, reflection separation, and robust perception under glare~\cite{wolff1997polarizationvision,atkinson2006diffusepolarization,nayar1997separation,serres2024passivepolarized}. Recent systems predict polarization directly from a conventional image~\cite{lin2025rgb2pol,zhang2025polaranything}, but their attainable fidelity is bounded by the labels used for supervision.

\begin{figure}[t]
  \centering
  \includegraphics[height=0.26\textheight]{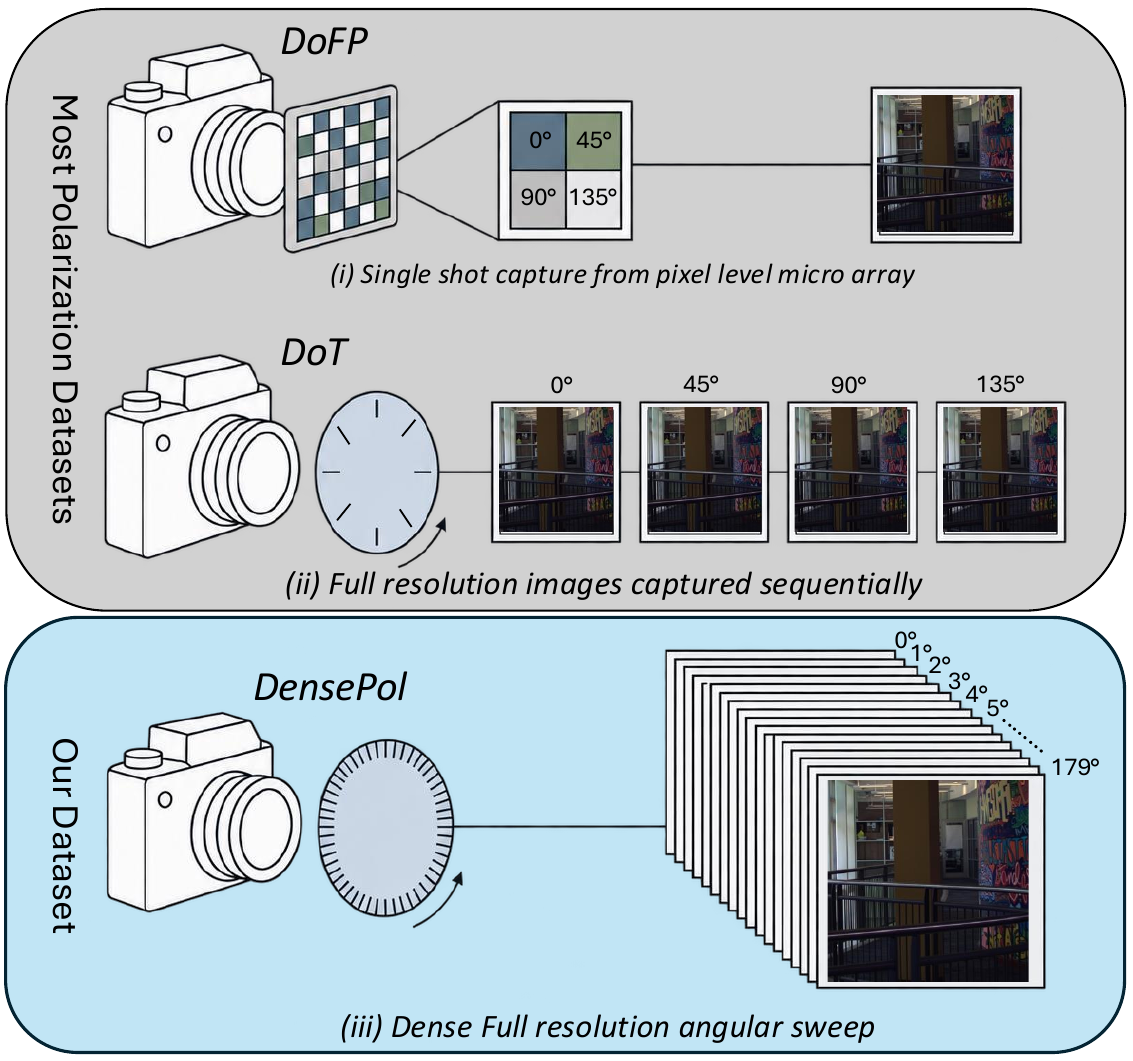}
  \caption{\textbf{Dataset Capture Process} Most polarization datasets are captured through Division of Focal Plane (i) or Division of Time (ii) approaches. DensePol (iii) increases polarization accuracy through a dense angular sweep from a rotating analyzer at $1\dg$ intervals. }
  \label{fig:dataset_differences}
  \vspace{-14pt}
\end{figure}

Most practical datasets use Division-of-Focal-Plane (DoFP) cameras. Figure~\ref{fig:dataset_differences} contrasts the two standard acquisition modes with ours. Their four analyzer orientations are spatially interleaved, so reconstructed Stokes maps inherit micro-polarizer calibration, interpolation, and instantaneous-field-of-view errors~\cite{powell2013calibrationdofp,zhang2016interpolationdofp,ratliff2009ifov}. Four measurements suffice for the ideal three-parameter linear-Stokes model but leave only one residual constraint. Structured sensor and reconstruction errors can consequently become part of the target learned by an RGB-to-polarization model.

For static scenes, Division-of-Time (DoT) acquisition instead rotates one full-resolution analyzer~\cite{tyo2006review,riviere2017reflectometry}. Dense angular scanning itself is not new: Perkins and Gruev swept a uniform polarized calibration source through $180\dg$ in $1\dg$ increments, with repeated frames at each angle and intensity, to validate a DoFP Stokes-noise model~\cite{perkins2010snr}. Their reported measurements are a controlled sensor-calibration experiment. For scene acquisition, the value of dense sampling is overdetermination, noise averaging, and a residual spectrum that exposes departures from the ideal angular model. DoT remains vulnerable to scene motion, mechanical error, and systematics at the retained harmonic, so a dense sweep is a high-redundancy reference rather than proof of absolute accuracy.

We introduce \method, containing $2{,}018$ paired RGB--polarization images where polarization comes from $180$ angles. We retain the angular stacks, commanded-angle metadata, and harmonic residuals rather than distributing only derived maps. We use the data to study angular-count stability and adapt deterministic diffusion-based dense prediction to single-RGB polarization synthesis. The two-stage model predicts bounded DoLP and a circular doubled-angle representation and supports a compact component-preserving refiner.

Figure~\ref{fig:overview} summarizes \method's two roles: dense DoT acquisition provides a high-redundancy polarization reference, and the resulting RGB--polarization pairs supervise single-image polarization prediction.
Our contributions are:
\begin{enumerate}
  \item \textbf{A dense-angle RGB--polarization dataset:} $2{,}018$ complete $180$-view polarization stacks, from indoor, outdoor, and synthetic scenes. 
  \item \textbf{Sampling, residual, and learning studies:} capture-level convergence, disjoint-subset agreement, held-out angular prediction, and controlled label-budget training, with explicit limits on treating the $180$-view fit as ground truth.
  \item \textbf{Deterministic RGB-to-polarization synthesis:} A diffusion-based predictor with circular output, native-resolution inference, and a DoLP refiner network that improves DoLP magnitude and structure without rotating AoLP.
  
\end{enumerate}

\section{Related Work}
\label{sec:related}

\paragraph{Applications of Polarimetric vision}
Polarization provides complementary cues for surface orientation, material identity, and reflection separation~\cite{atkinson2006diffusepolarization,wolff1990material,nayar1997separation}. These cues support learned shape recovery~\cite{ba2020deepsfp}, road-scene analysis~\cite{blin2020polarroads}, material segmentation~\cite{liang2022mcubes}, and glass segmentation~\cite{mei2022rgbpglass}. Such applications generally assume polarimetric measurements at inference; RGB-to-polarization synthesis instead seeks to recover useful polarization cues from conventional imagery, making prediction quality dependent on the fidelity of the acquisition targets.

\vspace{-12pt}

\paragraph{Polarimetric acquisition}
DoFP sensors trade spatial/angular fidelity for single-shot capture, with well-characterized calibration and Instantaneous Field of View (IFOV) artifacts~\cite{powell2013calibrationdofp,gruev2010ccd,zhang2016interpolationdofp,ratliff2009ifov}; DoT systems preserve full spatial sampling for static scenes at the cost of temporal and mechanical sensitivity~\cite{tyo2006review,riviere2017reflectometry}. Prior non-mosaiced reference datasets and their resolution, registration, and sharpness limitations are reviewed by Bigu\'e et al.~\cite{bigue2023reference}. \method targets this complementary regime and retains all measurements needed to inspect both the fitted polarization state and its residual.
\vspace{-12pt}
\paragraph{RGB-to-polarization learning}
PolarAnything adapts a pretrained diffusion backbone to synthesize polarization from one RGB image~\cite{zhang2025polaranything}. GenPolar uses a Stokes-informed two-stage diffusion formulation that receives fitted RGB $S_0$ and estimates absolute $S_1, S_2$~\cite{luo2026genpolar}, while Lin et al.~\cite{lin2025rgb2pol} benchmark RGB-to-polarization estimation with restoration and transformer architectures. Latent diffusion supplies a strong visual prior~\cite{rombach2022latentdiffusion}; $\mathrm{D}^{3}$-Predictor converts that prior into deterministic dense prediction through feature alignment between clean and noisy branches~\cite{xia2025d3predictor}. We adapt the $\mathrm{D}^{3}$-Predictor formulation to dense-DoT labels, encode AoLP cyclically, and evaluate arbitrary image sizes without global resizing. We retrain Restormer~\cite{zamir2022restormer}, Uformer~\cite{wang2022uformer}, and MAE~\cite{he2022mae} on the training partition of the same split as feed-forward references.

\section{The \method Dataset}
\label{sec:dataset}

\method is a paired RGB--polarization dataset built to provide a highly redundant reference for linear-polarization supervision. This section describes the dense DoT acquisition (Sec.~\ref{sec:acq}), the harmonic estimation and residual analysis that turn $180$ raw frames into Stokes reference maps (Sec.~\ref{sec:fdgt}), and the composition and encoding of the released data (Sec.~\ref{sec:composition}).

\subsection{Dense Division-of-Time Acquisition}
\label{sec:acq}


The two raw-backed collections comprise full-frame tabletop-scene and multi-pose-object captures. For each static scene, an IDS U3-3990SE-C-HQ color area-scan camera~\cite{idsU33990SE} acquired a mosaic over a fixed $2500\times2500$ sensor region through a linear polarizer mounted in a Thorlabs ELL14 Elliptec motorized rotation mount~\cite{thorlabsELL14}. We captured one frame at each of $180$ commanded analyzer orientations, from $0\dg$ to $179\dg$ in $1\dg$ increments, thereby sampling one complete period of linear polarization at full spatial resolution. %

%
%
\begin{table}[t]
  \centering
  \footnotesize
  \label{tab:dataset_comparison}
  \setlength{\tabcolsep}{1.0pt}
  \renewcommand{\arraystretch}{1.14}
  \begin{tabular}{@{}lcccc@{}}
    \toprule
    Dataset & Samples & Resolution & Acq. & Angles \\
    \midrule
    Morimatsu et al.~\cite{morimatsu2020eari}
      & $40$ & $1024{\times}768$ & DoT & $4$ \\
    Qiu et al.~\cite{qiu2019polarizationdemosaicking}
      & $40$ & $1024{\times}1024$ & DoT & $4$ \\
    Wen et al.~\cite{wen2021sparsejoint}
      & $50$ & $540{\times}720$ & DoT & $4$ \\
    Sparse-PDM~\cite{liu2023sparsepdm}
      & $300$ & $2448{\times}2048$ & DoT/DoFP & $4$ \\
    Kurita et al.~\cite{kurita2023sparsepolarization}
      & $1{,}049$ & $5.0$/$20.0$\,MP & DoFP/DoT & $4$ \\
    Jeon et al.~\cite{jeon2024spectral}
      & $2{,}022$ & $1900{\times}2100$ & DoFP & $4/8$ \\
    PolaRGB~\cite{yao2025polarfree}
      & $6{,}500$ & $1224{\times}1024$ & DoFP & $4$ \\
    PIDSR~\cite{zhou2025pidsr}
      & $138$ & $\leq2048{\times}2448$ & DoT & $4$ \\
    Abdul Rahman et al.~\cite{rahman2025polarizationdenoising}
      & $120$ & $1024{\times}768$ & DoT & $4$ \\
    PolarNS~\cite{hwang2025polarburst}
      & $244$ & $2448{\times}2048$ & DoFP & $4$ \\
    PolarBurstSR~\cite{hwang2025polarburst}
      & $160$ & $2448{\times}2048$ & DoFP & $4$ \\
    PolarAnything~\cite{zhang2025polaranything}
      & $1{,}148$ & $1224{\times}1024$ & DoFP & $4$ \\
    \midrule
    \textbf{\method (ours)}
      & \textbf{2,018} & $\mathbf{1250{\times}1250}$ & \textbf{DoT} & $\mathbf{180}$ \\
    \bottomrule
    
  \end{tabular}
  \caption{\textbf{Selected color--polarization datasets.} Comparison of dataset size, resolution, acquisition type, and number of analyzer angles.}
  \vspace{-12pt}
\end{table}
After each polarization sweep, we removed the polarizer without moving the camera or scene and captured a separate image from the same sensor region. Automatic exposure, gain, and white balance were briefly enabled and then fixed before capture. Thus, the paired RGB is a distinct polarizer-removed observation rather than an analyzer view or an image derived from $S_0$. We partition each full-resolution RGB--polarization pair into four non-overlapping $1250\times1250$ quadrants for release and learning, while performing angular analyses on the original full-frame sweep. Compared with four-angle DoFP acquisition, this protocol provides $45\times$ denser angular sampling while preserving full spatial sampling.

\subsection{Harmonic Estimation and Residual Analysis}
\label{sec:fdgt}

Under ideal linear polarimetry, the intensity measured at analyzer angle $\theta$ follows generalized Malus' law for an ideal linear analyzer~\cite{adams2019optics},
\begin{equation}
  I(\theta) = \tfrac{1}{2}\, S_0 \big( 1 + \rho \cos\!\big(2(\theta-\phi)\big) \big),
  \label{eq:malus}
\end{equation}
where $\rho$ is the degree of linear polarization (DoLP) and $\phi$ the angle of linear polarization (AoLP). Equivalently, the 
angular signal consists of a DC component and the physical $2\theta$ harmonic:
\begin{equation}
  I(\theta) = a_0 + a_2 \cos 2\theta + b_2 \sin 2\theta .
  \label{eq:harmonic}
\end{equation}
where $a_0$, $a_2$, and $b_2$ are constants.
A four-angle measurement at $0\dg/45\dg/90\dg/135\dg$ suffices to estimate these three coefficients 
under the ideal model, but provides only one residual degree of freedom,
$I_0+I_{90}=I_{45}+I_{135}$. This constraint can detect some model violations, but cannot localize a corrupted measurement or distinguish higher-order angular components that alias onto the retained DC and $2\theta$ terms.


With $N=180$ uniformly sampled angles indexed by $k\in\{0,\ldots,N-1\}$, $\theta_k=k\pi/N$, we estimate the same three-parameter model by overdetermined least squares.
Let $\mathbf{i}=[I(\theta_0),\ldots,I(\theta_{N-1})]^\top$ and let $\mathbf{X}$ contain columns $[1,\cos 2\theta_k,\sin 2\theta_k]$. The estimate is
\begin{equation}
  \widehat{\boldsymbol{\beta}}
  =(\mathbf{X}^{\top}\mathbf{X})^{-1}\mathbf{X}^{\top}\mathbf{i},
  \qquad
  \boldsymbol{\beta}=[a_0,a_2,b_2]^{\top}.
  \label{eq:harmonic_ls}
\end{equation}
Because the angles uniformly sample a complete $180\dg$ period, these basis columns are orthogonal. Consequently, Eq.~\eqref{eq:harmonic_ls} is equivalent to extracting the DC and first non-DC Fourier coefficients corresponding to the physical $2\theta$ harmonic from the discrete Fourier transform. We use the DFT as an efficient implementation of this least-squares projection; for nonuniform analyzer angles, Eq.~\eqref{eq:harmonic_ls} is instead evaluated using the corresponding angles.
The Stokes parameters are then
\begin{equation}
  S_0 = 2a_0,\quad S_1 = 2a_2,\quad S_2 = 2b_2,
  \label{eq:stokes}
\end{equation}
from which we derive
\begin{equation}
  \mathrm{DoLP} = \frac{\sqrt{S_1^2 + S_2^2}}{S_0},\qquad
  \mathrm{AoLP} = \tfrac{1}{2}\,\mathrm{atan2}(S_2, S_1).
  \label{eq:dolp_aolp}
\end{equation}
We demosaic each raw frame with the verified RGB convention before fitting these coefficients per color channel. The released scalar DoLP/AoLP supervision combines the full color polarization fit into the dense polarization reference used by all \method reconstruction metrics. 
The $180$ samples provide $177$ residual degrees of freedom, 
reduce variance from independent measurement noise, and expose non-model angular structure that can alias into a four-angle estimate.
\begin{table}[t]
  \centering
  \small
  \begin{tabular}{lcccc}
    \toprule
    Subset & Train & Val & Test & Total \\
    \midrule
    Tabletop scenes & $830$ & $108$ & $93$ & $1{,}031$ \\
    Multi-pose objects & $608$ & $70$ & $75$ & $753$ \\
    Outdoor & $130$ & $45$ & $29$ & $204$ \\
    Synthetic & $24$ & $3$ & $3$ & $30$ \\
    \midrule
    \textbf{Total images} & $\mathbf{1{,}592}$ & $\mathbf{226}$ & $\mathbf{200}$ & $\mathbf{2{,}018}$ \\
    \bottomrule
  \end{tabular}
  \caption{\method split and subset rows report image pairs.}
  \label{tab:dataset}
  \vspace{-12pt}
\end{table}
Under independent, equal-variance frame noise, the coefficient variance scales as $1/N$. Consequently, using $180$ rather than four analyzer angles reduces the variance by $45\times$, corresponding to an approximately $\sqrt{45}\approx6.7\times$ reduction in standard deviation. The least-squares fit retains the DC and $2\theta$ components prescribed by the ideal linear-polarization model, while the residual angular spectrum provides a per-pixel diagnostic of model mismatch. We therefore use the resulting $180$-view Stokes maps as the measurement-derived ground truth for training and evaluation. Sec.~\ref{sec:angle_convergence} examines how the fitted polarization state stabilizes as the angular sample count increases.

\subsection{Composition}
\label{sec:composition}

\method contains $2{,}018$ RGB--polarization pairs as seen in Table~\ref{tab:dataset}, which comprises of $1{,}988$ real pairs and $30$ physically based polarized renderings generated with Mitsuba~3~\cite{nimierdavid2019mitsuba2}. Of the real pairs, $1{,}784$ come from the two raw-backed collections: $1{,}031$ tabletop-scene and $753$ multi-pose-object pairs. The remaining $204$ are outdoor pairs, each traceable to a complete retained $180$-frame angular stack. To prevent cross-split leakage, we assign acquisition groups exclusively to one partition and audit the resulting split using exact pixel hashes, perceptual hashes, and DINOv2~\cite{oquab2023dinov2} feature similarity.

\begin{figure*}[t]
  \centering
  \includegraphics[width=0.98\linewidth]{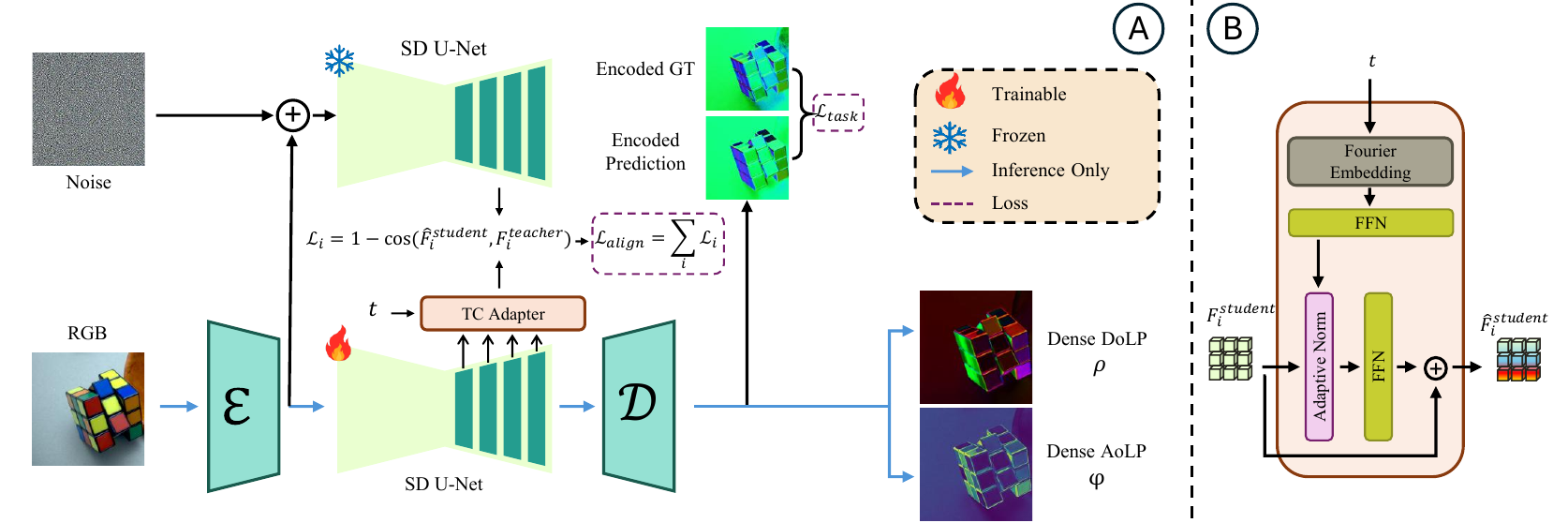}
  \caption{\textbf{Overview of the Stage I diffusion predictor.} (A) A trainable SD U-Net predicts the encoded polarization representation from RGB, while a frozen SD expert provides multi-scale features from noise-corrupted latents. Timestep-conditioned adapters align the student and expert features through \(\mathcal{L}_{\mathrm{align}}\), together with the task supervision \(\mathcal{L}_{\mathrm{task}}\); only the prediction branch is retained at inference. (B) The timestep-conditioned (TC) adapter embeds the diffusion timestep and uses it to modulate the student feature through adaptive normalization and an FFN with residual refinement.}
  \label{fig:stage1}
  \vspace{-12pt}
\end{figure*}

\section{Method}
\label{sec:method}

This section formulates the single-image polarization estimation problem and introduces the polarization representation used throughout the network. We then present the proposed two-stage framework and its training objectives.

\subsection{Problem Formulation}
\label{sec:problem}

Given a single RGB image $I \in \mathbb{R}^{3\times H\times W}$, our goal is to estimate its per-pixel linear polarization state, represented by the Degree of Linear Polarization (DoLP) $\rho$ and Angle of Linear Polarization (AoLP) $\phi$. Since AoLP is periodic over $[0,\pi)$, directly regressing $\phi$ introduces a discontinuity at the angular boundary. We therefore represent the polarization map using a three-channel circular encoding:
\begin{equation}
    P =
    \big[
    \rho,\,
    \cos(2\phi),\,
    \sin(2\phi)
    \big].
\end{equation}

The proposed model predicts an encoded polarization map
$\hat{P}\in\mathbb{R}^{3\times H\times W}$, where the first channel corresponds to DoLP and the remaining two channels represent AoLP in the double-angle space. The angular components are normalized to unit magnitude, and AoLP is recovered as
\begin{equation}
    \hat{\phi}
    =
    \frac{1}{2}
    \operatorname{atan2}
    \left(
    \hat{P}_{3},
    \hat{P}_{2}
    \right).
\end{equation}

\subsection{Network Overview}
\label{sec:overview}

The proposed framework consists of two sequential stages for estimating polarization from a single RGB image $I$. In the first stage, shown in Fig.~\ref{fig:stage1}, a diffusion-based predictor estimates a coarse encoded polarization map
$P_{\mathrm{enc}}=[\rho,\cos(2\phi),\sin(2\phi)]$.

In the second stage, shown in Fig.~\ref{fig:stage2}, the coarse prediction is combined with the original RGB image and processed by a lightweight local refinement network. The refiner predicts a gated residual correction for the DoLP component, producing the final estimate $\hat{\rho}$ while preserving the angular prediction from Stage I.

\subsection{Stage I: Diffusion-Based Polarization Prediction}
\label{sec:stage1_method}

As shown in Fig.~\ref{fig:stage1}, Stage I leverages the representation learned by a pretrained Stable Diffusion model to predict polarization directly from a single RGB image. Unlike conventional diffusion generation, our predictor operates in a single forward pass without iterative denoising. The RGB image is first encoded into the latent space using the pretrained VAE, processed by a trainable diffusion U-Net at a learned timestep, and decoded to obtain the coarse polarization prediction $P_{\mathrm{enc}}$.

\noindent
\textit{One-Step Polarization Prediction.}
Given the input image $I$, the frozen VAE encoder produces a latent representation
\begin{equation}
    z = \mathcal{E}_{\mathrm{VAE}}(I).
\end{equation}
The latent is then processed by the polarization predictor $\mathcal{D}_{\theta}$ at a learnable diffusion timestep $\tau$:
\begin{equation}
    \hat{z}_{P} = \mathcal{D}_{\theta}(z,\tau,c),
\end{equation}
where $c$ denotes the fixed text conditioning. Finally, the VAE decoder maps the predicted latent back to the image domain,
\begin{equation}
    P_{\mathrm{enc}}
    =
    \mathcal{D}_{\mathrm{VAE}}(\hat{z}_{P}),
\end{equation}
yielding the three-channel coarse polarization representation introduced in Sec.~\ref{sec:problem}.

\noindent
\textit{Multi-Timestep Diffusion Feature Alignment.}
Following the diffusion feature alignment strategy of D3-Predictor~\cite{xia2025d3predictor}, we use a second SD2.1 U-Net~\cite{rombach2022latentdiffusion} as a frozen teacher to guide the one-step polarization predictor during training. Both the student and teacher U-Nets are initialized from the same pretrained SD2.1 weights, while only the student polarization predictor is optimized. For each clean latent $z$, we sample multiple diffusion timesteps $\{t_k\}$ across the diffusion trajectory and generate the corresponding noisy latents:

\begin{equation}
    z_{t_k}
    =
    \sqrt{\bar{\alpha}_{t_k}}\,z
    +
    \sqrt{1-\bar{\alpha}_{t_k}}\,\epsilon,
\end{equation}
where $\epsilon \sim \mathcal{N}(0,I)$.

The noisy latents are processed by the frozen teacher to obtain intermediate features $F_i^{\mathrm{teacher}}(t_k)$ at feature level $i$. In parallel, the student processes the clean latent only once at the learned timestep $\tau$, producing features $F_i^{\mathrm{student}}$. Since the teacher features correspond to different diffusion timesteps, we introduce a \emph{Timestep-Conditioned Adapter (TC-Adapter)} $A_i$ that transforms each student feature according to the target timestep:
\begin{equation}
    \hat{F}_i^{\mathrm{student}}(t_k)
    =
    A_i\!\left(F_i^{\mathrm{student}},t_k\right).
\end{equation}

The adapted student features are then encouraged to match the corresponding teacher features using cosine similarity:
\begin{equation}
    \mathcal{L}_{\mathrm{align}}
    =
    \sum_{i,k}
    \left[
    1-
    \operatorname{cos}
    \left(
    \hat{F}_i^{\mathrm{student}}(t_k),
    F_i^{\mathrm{teacher}}(t_k)
    \right)
    \right].
\end{equation}
This allows the single-step student to learn representations associated with multiple stages of the pretrained diffusion process without performing iterative denoising.

\subsection{Stage II: Local DoLP Refinement}
\label{sec:stage2_method}

\begin{figure}[h]
    \centering
    \includegraphics[
        page=2,
        width=\linewidth
    ]{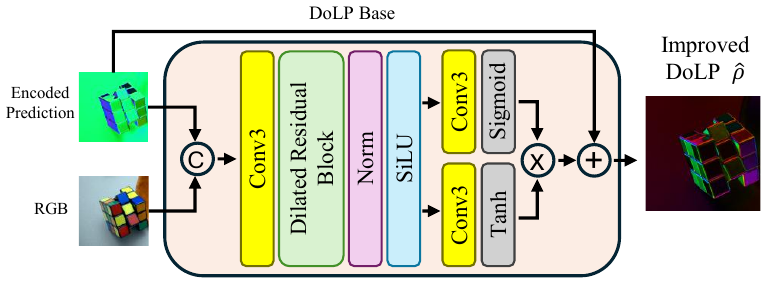}
    \caption{\textbf{Stage II local DoLP refiner.} RGB and the Stage I encoded prediction are fused to estimate a gated residual correction for refining DoLP.}
    \label{fig:stage2}
\end{figure}

Although Stage I provides a strong polarization estimate, its DoLP prediction can still miss fine local structures. As shown in Fig.~\ref{fig:stage2}, Stage II introduces a lightweight full-resolution refinement network that combines the original RGB image $I$ with the encoded polarization prediction $P_{\mathrm{enc}}$ to recover finer DoLP details. Rather than re-estimating polarization from scratch, the network learns a residual correction to the Stage I DoLP prediction.

\noindent
\textit{RGB--Polarization Feature Fusion.}
The RGB image and encoded polarization prediction are concatenated along the channel dimension and projected into a shared feature space using a $3\times3$ convolution:
\begin{equation}
    F_0 = \mathrm{Conv}_{3\times3}\!\left([I,P_{\mathrm{enc}}]\right).
\end{equation}
The fused features are then processed by a sequence of full-resolution residual blocks with varying dilation rates, allowing the network to capture local context at different receptive fields without spatial downsampling.
\noindent
\textit{Gated Residual Correction.}
From the refined features $F$, the network predicts a residual $\Delta\rho$ and a spatial gate $G$ to selectively correct the Stage-I DoLP estimate. The final DoLP is obtained as
\begin{equation}
    \hat{\rho}
    =
    \rho_{\mathrm{base}}
    +
    \alpha\,G\odot\Delta\rho,
\end{equation}
where $\alpha$ controls the residual magnitude. The AoLP prediction is retained from Stage I.

\subsection{Loss Function}
\label{sec:loss}

Stage I is optimized using a polarization reconstruction objective together with the diffusion feature-alignment loss introduced above. The task loss combines encoded-map reconstruction, DoLP consistency, circular AoLP consistency, angular-vector regularization, and spatial gradient preservation:
\begin{equation}
    \mathcal{L}_{\mathrm{task}}
    =
    \mathcal{L}_{\mathrm{rec}}
    +
    \mathcal{L}_{\mathrm{DoLP}}
    +
    \mathcal{L}_{\mathrm{AoLP}}
    +
    \mathcal{L}_{\mathrm{unit}}
    +
    \mathcal{L}_{\mathrm{grad}}.
\end{equation}
The AoLP term is computed in the double-angle representation to avoid discontinuities caused by angular periodicity. The overall Stage-I objective is
\begin{equation}
    \mathcal{L}_{\mathrm{Stage\,I}}
    =
    \mathcal{L}_{\mathrm{task}}
    +
    \mathcal{L}_{\mathrm{align}}.
\end{equation}

Stage II is trained separately to refine the DoLP prediction by minimizing the discrepancy between the refined estimate $\hat{\rho}$ and its ground-truth DoLP $\rho_{\mathrm{gt}}$, denoted by $\mathcal{L}_{\mathrm{ref}}$. This allows the refinement network to focus specifically on correcting local DoLP errors while retaining the Stage-I angular prediction.

\section{Experiments}
\label{sec:exp}

\noindent\textbf{Implementation Details.}
We implement our model in PyTorch and initialize Stage I from Stable Diffusion~2.1. Training uses random $512\times512$ crops, AdamW optimization, an effective batch size of four, gradient clipping, and a short warm-up followed by linear learning-rate decay. Stage I is trained for $150$k steps on a single NVIDIA H100 NVL GPU, requiring approximately $30.3$ GPU-hours, with checkpoint selection based only on validation performance. At inference, arbitrary-resolution images are processed using overlapping $1024\times1024$ tiles with $256$-pixel overlap. After Stage I is selected and frozen, the six-block local DoLP refiner is trained on $768\times768$ crops for up to $10$k steps while preserving the Stage-I AoLP prediction.

\noindent\textbf{Evaluation Metrics.}
We report image-averaged DoLP MAE, PSNR, and SSIM, together with circular AoLP MAE. AoLP is evaluated where reference DoLP $>0.05$. All metrics are computed on physically decoded DoLP and AoLP maps at native resolution, with the test set used only after model selection.

\subsection{Angular-Count Convergence}
\label{sec:angle_convergence}

We first ask whether increasing the analyzer-angle budget makes the fitted polarization state more stable. For $N\in\{4,8,16,32,90,180\}$ and $j\in\{0,\ldots,N-1\}$, we select approximately uniform samples $\lfloor 180j/N\rfloor$ and refit Eq.~\eqref{eq:harmonic_ls} at the corresponding recorded analyzer angles. We then evaluate every distinct integer phase rotation ($45$ for $N\leq32$ and two for $N=90$). We measure (i) deviation from the $180$-view reference in normalized Stokes $(q,u)=(S_1/S_0,S_2/S_0)$, DoLP, and circular AoLP; (ii) agreement between two non-overlapping equal-budget fits; and (iii) prediction error on analyzer frames excluded from the fit. The held-out normalized root-mean-square error (NRMSE) is computed per pixel as temporal RMSE divided by that pixel's dense mean intensity, then averaged over pixels.

We sample a stride-$32$ spatial lattice while retaining all four pixel parity classes, excluding pixels with dense mean $<5$ DN or any value $\geq250$ DN; AoLP additionally requires reference DoLP $>0.05$. We take the median over phase rotations per acquisition and average across acquisitions; $95\%$ CIs use cluster bootstrap with poses of the same physical object resampled jointly.


\begin{figure}[t]
  \centering
  \includegraphics[width=\linewidth]{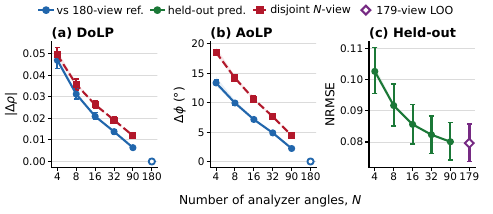}
  \caption{\textbf{Angular-count convergence.} Increasing the number of analyzer angles improves DoLP and AoLP stability and reduces held-out prediction error.}
  \label{fig:angle_count_convergence}
  \vspace{-8pt}
\end{figure}

Figure~\ref{fig:angle_count_convergence} shows monotonic DoLP, AoLP, and held-out improvements with angle count. From four to $90$ angles, the normalized-Stokes deviation falls from $0.0686$ to $0.0102$ ($85\%$), absolute DoLP deviation from $0.0469$ to $0.0064$ ($86\%$), and circular AoLP deviation from $13.36\dg$ (95\% CI: $12.89$--$13.81\dg$) to $2.21\dg$ ($2.07$--$2.35\dg$; $83\%$). Held-out intensity NRMSE decreases from $0.1026$ ($0.0955$--$0.1103$) to $0.0801$ ($0.0742$--$0.0863$), close to the $179$-view leave-one-out reference of $0.0796$ ($0.0738$--$0.0857$); $90\%$ of this predictive reduction is already achieved by $32$ views. The trend also holds separately for the tabletop-scene and multi-pose object collections, although the latter is harder: the four-to-$90$ AoLP deviations are $8.06\dg\!\to\!1.14\dg$ and $22.01\dg\!\to\!3.95\dg$, respectively. These results support improved internal stability and same-stack prediction with denser sampling.

\subsection{RGB-to-Polarization Synthesis}
\label{sec:rgb2pol_exp}
On the held-out $226$-image validation set, extending optimization is helpful through the middle of training but does not monotonically improve either polarization quantity. Masked AoLP reaches its minimum of $19.67\dg$ at $75$k steps; extending the same trajectory to $150$k yields $19.91\dg$. DoLP MAE is lowest among unrefined models at $100$k ($0.0511$), while the $75$k checkpoint differs by less than $0.001$. We therefore select $75$k for the primary angular field and do not claim that additional steps improve generalization.
\vspace{-12pt}
\begin{figure*}[t]
  \centering
  \includegraphics[width=0.97\linewidth]{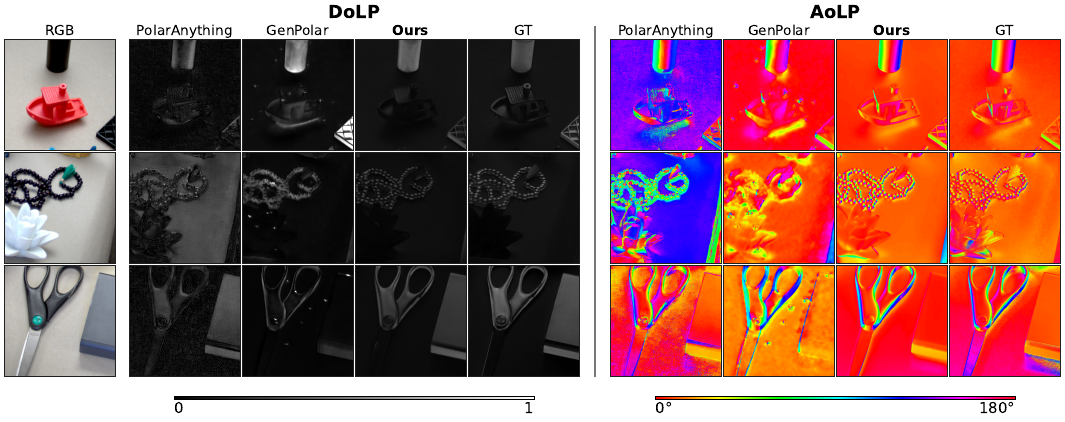}
 \caption{\textbf{Qualitative comparison.}
Results use the retrained models and inputs described in
Table~\ref{tab:rgb2pol_main}; \method includes the compact DoLP refiner.
GT denotes the reference. DoLP uses $[0,1]$ grayscale range,
and AoLP uses a cyclic $0$--$180\dg$ range.}
  \label{fig:rgb2pol_validation}
  \vspace{-12pt}
\end{figure*}
\paragraph{Reference angular density}
We next isolate how the angular density of the evaluation reference changes the measured error of one fixed predictor. We freeze the selected $75$k base model and score the same native-resolution test predictions against references fitted from $N\in\{4,8,16,32,90\}$ approximately uniform views drawn from the even-indexed half of each sweep. All five references use the same fitting and encoding path. For a row-comparable AoLP measure, every row uses one fixed DoLP $>0.05$ mask obtained from the dense, measurement-disjoint odd-indexed $90$-view fit; thus neither the evaluated reference nor its changing DoLP scale selects the scored pixels.

\begin{table}[t]
  \centering
  \scriptsize
  \resizebox{\columnwidth}{!}{%
  \begin{tabular}{@{}rcccc@{}}
    \toprule
    $N$ & AoLP$\downarrow$ & DoLP MAE$\downarrow$ & DoLP PSNR$\uparrow$ & DoLP SSIM$\uparrow$ \\
    \midrule
     $4$ & $22.15\dg$ & $0.0583$ & $23.00$ & $0.4849$ \\
     $8$ & $20.87\dg$ & $0.0531$ & $23.93$ & $0.5734$ \\
    $16$ & $20.20\dg$ & $0.0510$ & $24.36$ & $0.6345$ \\
    $32$ & $19.81\dg$ & $0.0501$ & $24.57$ & $0.6742$ \\
    $90$ & $\mathbf{19.55\dg}$ & $\mathbf{0.0495}$ & $\mathbf{24.70}$ & $\mathbf{0.7062}$ \\
    \bottomrule
  \end{tabular}}
  \caption{\textbf{Reference angular density.} Predictions from the same fixed model are evaluated against references fitted from different numbers of analyzer views.}
  \label{tab:reference_count_probe}
  \vspace{-12pt}
\end{table}

Table~\ref{tab:reference_count_probe} shows monotonic improvement as the reference becomes denser. From four to $90$ views, common-mask AoLP MAE falls by $2.60\dg$ ($22.15\dg\!\to\!19.55\dg$), DoLP MAE falls by $15\%$ ($0.0583\!\to\!0.0495$), PSNR rises by $1.70$ dB, and SSIM rises by $0.221$. The disjoint even- and odd-indexed $90$-view references give nearly identical AoLP errors ($19.55\dg$ and $19.54\dg$), indicating that the curve is close to convergence. This experiment quantifies sensitivity to the evaluation reference.
\vspace{-12pt}
\paragraph{Training-label angular density}
A complementary single-seed experiment varies the angular density of the \emph{training} labels while holding inputs, split, initialization, and optimization fixed. At $75$k steps, masked AoLP against the common released reference changes from $20.05\dg$ at four views to $19.35\dg$ at $90$, while DoLP is best at $N=16$.

\vspace{-12pt}

\paragraph{Domain behavior}
At $75$k, masked AoLP error is $12.50\dg$ on tabletop scenes, $24.95\dg$ on multi-pose objects, $28.04\dg$ outdoors, and $29.33\dg$ on synthetic validation images; corresponding DoLP MAEs are $0.0524/0.0511/0.0523/0.0628$. Thus the similar aggregate DoLP errors conceal a pronounced angular domain gap. A controlled ablation trained and evaluated without the multi-pose object subset selects $75$k at $17.43\dg$ on the remaining $156$ validation images, but the all-domain model evaluated on those same images is slightly better in AoLP ($17.30\dg$). Removing that subset therefore does not solve angular generalization; it mainly improves DoLP on the remaining domains ($0.0525\!\to\!0.0498$).
\vspace{-12pt}
\paragraph{Refinement and saturation}
The selected six-block DoLP specialist lowers MAE by $7.4\%$ ($0.0521\!\to\!0.0482$), raises PSNR by $0.63$ dB, and raises SSIM by $0.141$ ($0.532\!\to\!0.673$). Because it copies the doubled-angle channels, AoLP is unchanged. A separately selected cyclic AoLP specialist changes aggregate masked error by only $0.004\dg$; the cascade gain therefore comes from magnitude/structure refinement.
\begin{figure*}[t]
  \centering
  \includegraphics[width=0.9\linewidth]{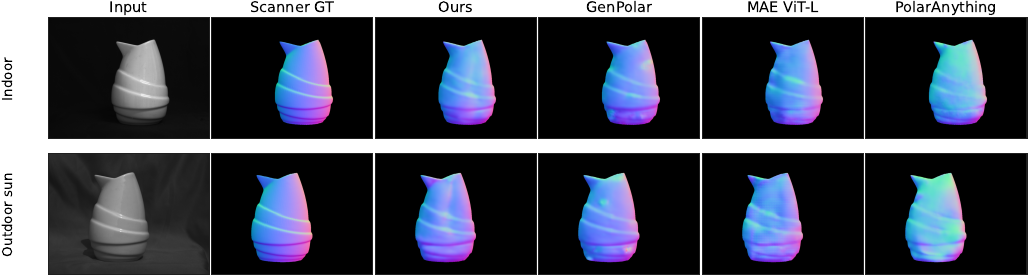}
  \caption{\textbf{Surface-normal transfer.} Representative indoor and outdoor DeepSfP results using the common frozen SfPNet and the models trained or retrained on DensePol, as described in Table~\ref{tab:downstream_normals}. Normal color encodes $(n_x,n_y,n_z)$ as RGB; black denotes background outside the scanner mask.}
\label{fig:deepsfp_retrained_normals}
\vspace{-12pt}
\end{figure*}

\vspace{-12pt}
\paragraph{Retrained reference}
Every baseline is retrained on the \method training partition with checkpoint selection restricted to validation. Table~\ref{tab:rgb2pol_main} gives the resulting accuracy on our test partition and Figure~\ref{fig:rgb2pol_validation} the corresponding qualitative comparison. Despite receiving only polarizer-removed RGB, \method lowers AoLP error by $13.78\dg$ relative to the method-faithful GenPolar row, which retains its privileged fitted-$S_0$ input, and by $11.15\dg$ relative to the strongest feed-forward baseline. It also improves DoLP PSNR by $4.81$ dB relative to GenPolar.
\vspace{-12pt}

\paragraph{Downstream surface-normal transfer}
We test whether synthesized polarization preserves cues beyond direct matching on the DeepSfP split~\cite{ba2020deepsfp}. One normal network is trained once on $236$ measured-polarization captures and frozen; only its polarization input changes on $21$ captures from seven unseen objects. Because DeepSfP has no color RGB, each generator receives the four-analyzer mean repeated over three channels and is evaluated at native resolution before the official central crop. Table~\ref{tab:downstream_normals} is thus a grayscale/domain-shift transfer test, not a general color-image claim. Figure~\ref{fig:deepsfp_retrained_normals} shows representative normal maps for the same held-out object indoors and in sunlight. The reference is sparse and its DoLP column uses four-angle DoFP measurements rather than dense angles; Sec.~\ref{sec:angle_convergence} finds a $13.36\dg$ four-to-$180$-view discrepancy. Only the normal columns use independent $3$D-scanner geometry. Training exposure also differs: $97.3\%$ of PolarAnything's released training images share this sensor geometry and $46\%$ of GenPolar's corpus is DoFP, whereas \method has no DoFP training exposure.

\begin{table}[t]
  \centering
  \footnotesize
  \setlength{\tabcolsep}{2pt}
  \renewcommand{\arraystretch}{1.1}
  \resizebox{\columnwidth}{!}{%
  \begin{tabular}{@{}lccccc@{}}
    \toprule
    & \multicolumn{3}{c}{Surface normals (scanner truth)} & \multicolumn{2}{c}{DoFP ref.} \\
    \cmidrule(lr){2-4}\cmidrule(lr){5-6}
    Polarization source & Mean$\downarrow$ & $\leq10\dg\uparrow$ & $\leq30\dg\uparrow$ & AoLP$\downarrow$ & DoLP$\downarrow$ \\
    \midrule
    \textit{Measured polarization} & \textit{17.06} & \textit{34.8} & \textit{85.6} & --- & --- \\
    \midrule
    \method (ours) & $\mathbf{23.40}$ & $\mathbf{17.3}$ & $73.3$ & $\mathbf{36.86}$ & $\mathbf{0.0352}$ \\
    GenPolar~\cite{luo2026genpolar} & $23.52$ & $17.1$ & $\mathbf{73.7}$ & $40.68$ & $0.0406$ \\
    MAE ViT-L~\cite{he2022mae} & $24.80$ & $13.9$ & $70.2$ & $39.56$ & $0.0423$ \\
    Uformer-T~\cite{wang2022uformer} & $25.26$ & $14.7$ & $69.2$ & $40.40$ & $0.0433$ \\
    Restormer~\cite{zamir2022restormer} & $25.80$ & $14.0$ & $66.3$ & $39.35$ & $0.0396$ \\
    PolarAnything~\cite{zhang2025polaranything} & $30.67$ & $8.5$ & $51.5$ & $46.49$ & $0.1493$ \\
    \bottomrule
  \end{tabular}}
  \caption{\textbf{DeepSfP polarization-to-normal transfer} on $21$ object-disjoint test captures using a common frozen normal estimator. Normal metrics use scanner ground truth.}
  \label{tab:downstream_normals}
  \vspace{-12pt}
\end{table}

\method ranks first among six predictions in mean normal error ($23.40\dg$), with GenPolar following at $23.52\dg$ and retrained PolarAnything last at $30.67\dg$. \method also has the lowest four-angle DoFP-reference AoLP error ($36.86\dg$), while PolarAnything has the highest ($46.49\dg$). However, the intermediate rankings differ, showing that agreement with the sparse reference does not fully predict downstream normal quality.

\begin{table}[!t]
  \centering
  \footnotesize
  \renewcommand{\arraystretch}{1.15}
  \begin{tabular*}{\columnwidth}{@{\extracolsep{\fill}}lccc@{}}
    \toprule
    Method & PSNR$\uparrow$ & SSIM$\uparrow$ & AoLP$\downarrow$ \\
    \midrule
    Restormer~\cite{zamir2022restormer}            & $22.64$ & $0.543$ & $30.86\dg$ \\
    Uformer-T~\cite{wang2022uformer}               & $22.20$ & $0.511$ & $31.48\dg$ \\
    MAE ViT-L~\cite{he2022mae}                     & $22.25$ & $0.509$ & $33.20\dg$ \\
    PolarAnything~\cite{zhang2025polaranything}    & $18.35$ & $0.214$ & $44.43\dg$ \\
    GenPolar~\cite{luo2026genpolar}                & $20.43$ & $0.458$ & $33.49\dg$ \\
    \midrule
    \method SD2.1 & $\mathbf{25.24}$ & $\mathbf{0.618}$ & $\mathbf{19.71\dg}$ \\
    \bottomrule
  \end{tabular*}
  \caption{\textbf{RGB-to-polarization accuracy on the DensePol test set.} Comparison with retrained feed-forward and diffusion-based baselines using DoLP PSNR/SSIM and circular AoLP error.}
  \label{tab:rgb2pol_main}
  \vspace{-12pt}
\end{table}

On our test partition, \method leads every reported predictor on all three quantities, lowering AoLP error by $11.15\dg$ against the strongest feed-forward baseline and by $13.78\dg$ against the method-faithful GenPolar row.

\begin{table}[t]
  \centering
  \scriptsize
  \resizebox{\columnwidth}{!}{%
  \begin{tabular}{@{}lccccc@{}}
    \toprule
    Target / loss & Step & AoLP$\downarrow$ &
    DoLP MAE$\downarrow$ & PSNR$\uparrow$ & SSIM$\uparrow$ \\
    \midrule
    Circular
      & $75$k & $19.671\dg$ & $\mathbf{0.05208}$ & $22.83$ & $0.5316$ \\
    Circular + $\rho$ weight
      & $75$k & $\mathbf{19.566\dg}$ & $0.05267$ & $22.79$ & $0.5237$ \\
    Normalized Stokes
      & $50$k & $20.044\dg$ & $0.05225$ &
        $\mathbf{22.87}$ & $\mathbf{0.5342}$ \\
    \bottomrule
  \end{tabular}}
  \caption{\textbf{Target-representation ablation.} Comparison of circular and normalized-Stokes targets on the validation set.}
  \label{tab:representation_ablation}
  \vspace{-12pt}
\end{table}

\subsection{Ablation}

We compare the circular target, a circular representation with DoLP-weighted angular losses, and
normalized Stokes under matched training and validation selection.
Table~\ref{tab:representation_ablation} shows that weighting the angular losses by DoLP
improves AoLP by only $0.105\dg$ while degrading all three DoLP
metrics. Normalized Stokes is $0.373\dg$ worse in AoLP than the
circular target, despite slightly higher PSNR and SSIM. We therefore
retain the unweighted circular representation as the best-balanced
choice given this single-seed comparison.



\section{Conclusion}
\label{sec:conclusion}

We introduced DensePol, an RGB--polarization dataset with 180 full-resolution analyzer measurements per capture, providing high-redundancy supervision for single-image polarization estimation. Unlike conventional four-angle acquisition, DensePol uses dense Division-of-Time measurements while preserving full spatial resolution. Building on this supervision, we introduced a two-stage RGB-to-polarization framework combining a deterministic diffusion-based predictor with multi-timestep feature alignment and a lightweight local DoLP refiner. Experiments show that dense angular sampling improves polarization-reference stability and reduces sensitivity to sparse-angle measurements, while our model outperforms the evaluated diffusion and feed-forward baselines. The local refiner further improves DoLP magnitude and structure without altering the predicted AoLP, and downstream surface-normal experiments show that the synthesized polarization preserves useful physical cues.


{
    \small
    \bibliographystyle{ieeenat_fullname}
    \bibliography{main}

@article{wolff1997polarizationvision,
  author  = {Wolff, Lawrence B.},
  title   = {Polarization vision: a new sensory approach to image understanding},
  journal = {Image and Vision Computing},
  volume  = {15},
  number  = {2},
  pages   = {81--93},
  year    = {1997}
}

@article{wolff1990material,
  author  = {Wolff, Lawrence B.},
  title   = {Polarization-based material classification from specular reflection},
  journal = {IEEE Trans. Pattern Anal. Mach. Intell.},
  volume  = {12},
  number  = {11},
  pages   = {1059--1071},
  year    = {1990}
}

@article{nayar1997separation,
  author  = {Nayar, Shree K. and Fang, Xi-Sheng and Boult, Terrance},
  title   = {Separation of reflection components using color and polarization},
  journal = {Int. J. Comput. Vis.},
  volume  = {21},
  number  = {3},
  pages   = {163--186},
  year    = {1997}
}

@article{atkinson2006diffusepolarization,
  author  = {Atkinson, Gary A. and Hancock, Edwin R.},
  title   = {Recovery of surface orientation from diffuse polarization},
  journal = {IEEE Trans. Image Process.},
  volume  = {15},
  number  = {6},
  pages   = {1653--1664},
  year    = {2006}
}

@article{serres2024passivepolarized,
  author  = {Serres, Julien R. and Lapray, Pierre-Jean and Viollet, St{\'e}phane and Kronland-Martinet, Thomas and Moutenet, Antoine and Morel, Olivier and Bigu{\'e}, Laurent},
  title   = {Passive Polarized Vision for Autonomous Vehicles: A Review},
  journal = {Sensors},
  volume  = {24},
  number  = {11},
  pages   = {3312},
  year    = {2024},
  doi     = {10.3390/s24113312}
}

@inproceedings{ba2020deepsfp,
  author    = {Ba, Yunhao and Gilbert, Alex and Wang, Franklin and Yang, Jinfa and Chen, Rui and Wang, Yiqin and Yan, Lei and Shi, Boxin and Kadambi, Achuta},
  title     = {Deep shape from polarization},
  booktitle = {Eur. Conf. Comput. Vis. (ECCV)},
  year      = {2020}
}

@inproceedings{lin2025rgb2pol,
  author    = {Lin, Beibei and Yuan, Zifeng and Chen, Tingting},
  title     = {{RGB}-to-Polarization Estimation: A New Task and Benchmark Study},
  booktitle = {Adv. Neural Inform. Process. Syst. (NeurIPS) Datasets and Benchmarks Track},
  volume    = {38},
  year      = {2025}
}

@inproceedings{zhang2025polaranything,
  author    = {Zhang, Kailong and Lyu, Youwei and Guo, Heng and Li, Si and Ma, Zhanyu and Shi, Boxin},
  title     = {{PolarAnything}: Diffusion-based Polarimetric Image Synthesis},
  booktitle = {Int. Conf. Comput. Vis. (ICCV)},
  pages     = {26466--26476},
  year      = {2025}
}

@inproceedings{rombach2022latentdiffusion,
  author    = {Rombach, Robin and Blattmann, Andreas and Lorenz, Dominik and Esser, Patrick and Ommer, Bj{\"o}rn},
  title     = {High-Resolution Image Synthesis with Latent Diffusion Models},
  booktitle = {IEEE Conf. Comput. Vis. Pattern Recog. (CVPR)},
  pages     = {10684--10695},
  year      = {2022}
}

@article{xia2025d3predictor,
  author  = {Xia, Changliang and Jia, Chengyou and Luo, Minnan and Dang, Zhuohang and Shen, Xin and Ping, Bowen},
  title   = {{$\mathrm{D}^{3}$-Predictor}: Noise-Free Deterministic Diffusion for Dense Prediction},
  journal = {arXiv preprint arXiv:2512.07062},
  year    = {2025}
}

@article{powell2013calibrationdofp,
  author  = {Powell, Samuel B. and Gruev, Viktor},
  title   = {Calibration methods for division-of-focal-plane polarimeters},
  journal = {Optics Express},
  volume  = {21},
  number  = {18},
  pages   = {21039--21055},
  year    = {2013}
}

@article{zhang2016interpolationdofp,
  author  = {Zhang, Junchao and Luo, Haibo and Hui, Bin and Chang, Zheng},
  title   = {Image interpolation for division of focal plane polarimeters with intensity correlation},
  journal = {Optics Express},
  volume  = {24},
  number  = {18},
  pages   = {20799--20807},
  year    = {2016}
}

@article{ratliff2009ifov,
  author  = {Ratliff, Bradley M. and LaCasse, Charles F. and Tyo, J. Scott},
  title   = {Interpolation strategies for reducing {IFOV} artifacts in microgrid polarimeter imagery},
  journal = {Optics Express},
  volume  = {17},
  number  = {11},
  pages   = {9112--9125},
  year    = {2009}
}

@article{gruev2010ccd,
  author  = {Gruev, Viktor and Perkins, Rob and York, Timothy},
  title   = {{CCD} polarization imaging sensor with aluminum nanowire optical filters},
  journal = {Optics Express},
  volume  = {18},
  number  = {18},
  pages   = {19087--19094},
  year    = {2010}
}

@article{tyo2006review,
  author  = {Tyo, J. Scott and Goldstein, Dennis L. and Chenault, David B. and Shaw, Joseph A.},
  title   = {Review of passive imaging polarimetry for remote sensing applications},
  journal = {Applied Optics},
  volume  = {45},
  number  = {22},
  pages   = {5453--5469},
  year    = {2006}
}

@article{riviere2017reflectometry,
  author  = {Rivi\`ere, J\'er\'emy and Reshetouski, Ilya and Filipi, Luka and Ghosh, Abhijeet},
  title   = {Polarization imaging reflectometry in the wild},
  journal = {ACM Trans. Graph.},
  volume  = {36},
  number  = {6},
  pages   = {1--14},
  year    = {2017}
}

@article{perkins2010snr,
  author  = {Perkins, Robert and Gruev, Viktor},
  title   = {Signal-to-noise analysis of {Stokes} parameters in division of focal plane polarimeters},
  journal = {Optics Express},
  volume  = {18},
  number  = {25},
  pages   = {25815--25824},
  year    = {2010},
  doi     = {10.1364/OE.18.025815}
}

@article{nimierdavid2019mitsuba2,
  author  = {Nimier-David, Merlin and Vicini, Delio and Zeltner, Tizian and Jakob, Wenzel},
  title   = {{Mitsuba 2}: A Retargetable Forward and Inverse Renderer},
  journal = {ACM Trans. Graph.},
  volume  = {38},
  number  = {6},
  pages   = {203:1--203:17},
  year    = {2019},
  doi     = {10.1145/3355089.3356498}
}

@inproceedings{zamir2022restormer,
  author    = {Zamir, Syed Waqas and Arora, Aditya and Khan, Salman and Hayat, Munawar and Khan, Fahad Shahbaz and Yang, Ming-Hsuan},
  title     = {{Restormer}: Efficient transformer for high-resolution image restoration},
  booktitle = {IEEE Conf. Comput. Vis. Pattern Recog. (CVPR)},
  year      = {2022}
}

@inproceedings{wang2022uformer,
  author    = {Wang, Zhendong and Cun, Xiaodong and Bao, Jianmin and Zhou, Wengang and Liu, Jianzhuang and Li, Houqiang},
  title     = {{Uformer}: A general {U}-shaped transformer for image restoration},
  booktitle = {IEEE Conf. Comput. Vis. Pattern Recog. (CVPR)},
  year      = {2022}
}

@inproceedings{he2022mae,
  author    = {He, Kaiming and Chen, Xinlei and Xie, Saining and Li, Yanghao and Doll\'ar, Piotr and Girshick, Ross},
  title     = {Masked autoencoders are scalable vision learners},
  booktitle = {IEEE Conf. Comput. Vis. Pattern Recog. (CVPR)},
  year      = {2022}
}

@article{wen2021sparsejoint,
  author    = {Wen, Sijia and Zheng, Yinqiang and Lu, Feng},
  title     = {A Sparse Representation Based Joint Demosaicing Method for Single-Chip Polarized Color Sensor},
  journal   = {IEEE Trans. Image Process.},
  volume    = {30},
  pages     = {4171--4182},
  year      = {2021}
}

@inproceedings{qiu2019polarizationdemosaicking,
  author    = {Qiu, Simeng and Fu, Qiang and Wang, Congli and Heidrich, Wolfgang},
  title     = {Polarization Demosaicking for Monochrome and Color Polarization Focal Plane Arrays},
  booktitle = {Vision, Modeling and Visualization},
  publisher = {The Eurographics Association},
  year      = {2019},
  doi       = {10.2312/vmv.20191325}
}

@inproceedings{morimatsu2020eari,
  author    = {Morimatsu, Miki and Monno, Yusuke and Tanaka, Masayuki and Okutomi, Masatoshi},
  title     = {Monochrome and Color Polarization Demosaicking Using Edge-Aware Residual Interpolation},
  booktitle = {IEEE Int. Conf. Image Process. (ICIP)},
  pages     = {2571--2575},
  year      = {2020},
  doi       = {10.1109/ICIP40778.2020.9191085}
}

@inproceedings{blin2020polarroads,
  author    = {Blin, Rachel and Ainouz, Samia and Canu, St{\'e}phane and Meriaudeau, Fabrice},
  title     = {A New Multimodal {RGB} and Polarimetric Image Dataset for Road Scenes Analysis},
  booktitle = {IEEE Conf. Comput. Vis. Pattern Recog. Workshops (CVPRW)},
  pages     = {216--217},
  year      = {2020}
}

@inproceedings{liang2022mcubes,
  author    = {Liang, Yupeng and Wakaki, Ryosuke and Nobuhara, Shohei and Nishino, Ko},
  title     = {Multimodal Material Segmentation},
  booktitle = {IEEE Conf. Comput. Vis. Pattern Recog. (CVPR)},
  pages     = {19800--19808},
  year      = {2022}
}

@inproceedings{mei2022rgbpglass,
  author    = {Mei, Haiyang and Dong, Bo and Dong, Wen and Yang, Jiaxi and Baek, Seung-Hwan and Heide, Felix and Peers, Pieter and Wei, Xiaopeng and Yang, Xin},
  title     = {Glass Segmentation Using Intensity and Spectral Polarization Cues},
  booktitle = {IEEE Conf. Comput. Vis. Pattern Recog. (CVPR)},
  pages     = {12622--12631},
  year      = {2022}
}

@inproceedings{kurita2023sparsepolarization,
  author    = {Kurita, Teppei and Kondo, Yuhi and Sun, Legong and Moriuchi, Yusuke},
  title     = {Simultaneous Acquisition of High Quality {RGB} Image and Polarization Information Using a Sparse Polarization Sensor},
  booktitle = {IEEE Winter Conf. Appl. Comput. Vis. (WACV)},
  pages     = {178--188},
  year      = {2023}
}

@article{liu2023sparsepdm,
  author  = {Liu, Ju and Duan, Jin and Hao, Youfei and Chen, Guangqiu and Zhang, Hao and Zheng, Yue},
  title   = {Polarization image demosaicing and {RGB} image enhancement for a color polarization sparse focal plane array},
  journal = {Optics Express},
  volume  = {31},
  number  = {14},
  pages   = {23475--23490},
  year    = {2023},
  doi     = {10.1364/OE.494836}
}

@inproceedings{bigue2023reference,
  author    = {Bigu{\'e}, Laurent and Foulonneau, Alban and Lapray, Pierre-Jean},
  title     = {Production of high-resolution reference polarization images from real world scenes},
  booktitle = {Polarization Science and Remote Sensing XI},
  volume    = {12690},
  pages     = {126900B},
  publisher = {SPIE},
  year      = {2023},
  doi       = {10.1117/12.2677421}
}

@inproceedings{jeon2024spectral,
  author    = {Jeon, Yujin and Choi, Eunsue and Kim, Youngchan and Moon, Yunseong and Omer, Khalid and Heide, Felix and Baek, Seung-Hwan},
  title     = {Spectral and Polarization Vision: Spectro-polarimetric Real-world Dataset},
  booktitle = {IEEE Conf. Comput. Vis. Pattern Recog. (CVPR)},
  pages     = {22098--22108},
  year      = {2024}
}

@inproceedings{yao2025polarfree,
  author    = {Yao, Mingde and Wang, Menglu and Tam, King-Man and Li, Lingen and Xue, Tianfan and Gu, Jinwei},
  title     = {{PolarFree}: Polarization-based Reflection-Free Imaging},
  booktitle = {IEEE Conf. Comput. Vis. Pattern Recog. (CVPR)},
  pages     = {10890--10899},
  year      = {2025}
}

@inproceedings{zhou2025pidsr,
  author    = {Zhou, Shuangfan and Zhou, Chu and Lyu, Youwei and Guo, Heng and Ma, Zhanyu and Shi, Boxin and Sato, Imari},
  title     = {{PIDSR}: Complementary Polarized Image Demosaicing and Super-Resolution},
  booktitle = {IEEE Conf. Comput. Vis. Pattern Recog. (CVPR)},
  pages     = {16081--16090},
  year      = {2025}
}

@inproceedings{hwang2025polarburst,
  author    = {Hwang, Inseung and Choi, Kiseok and Ha, Hyunho and Kim, Min H.},
  title     = {Benchmarking Burst Super-Resolution for Polarization Images: Noise Dataset and Analysis},
  booktitle = {Int. Conf. Comput. Vis. (ICCV)},
  pages     = {24899--24909},
  year      = {2025}
}

@inproceedings{rahman2025polarizationdenoising,
  author    = {Abdul Rahman, Muhamad Daniel Ariff Bin and Monno, Yusuke and Tanaka, Masayuki and Okutomi, Masatoshi},
  title     = {Polarization Denoising and Demosaicking: Dataset and Baseline Method},
  booktitle = {IEEE Int. Conf. Image Process. (ICIP)},
  year      = {2025}
}

@article{luo2026genpolar,
  author  = {Luo, Yidong and Li, Chenggong and Feng, Yuchao and Shi, Boxin and Zhang, Junchao and Yuan, Xin},
  title   = {Stokes-Informed Diffusion for Robust Linear Polarization Estimation},
  journal = {arXiv preprint arXiv:2607.21239},
  year    = {2026}
}

@misc{idsU33990SE,
  author       = {{IDS Imaging Development Systems GmbH}},
  title        = {{U3-3990SE-C-HQ} Color {USB3} Camera},
  howpublished = {Product documentation},
  year         = {n.d.},
  url          = {https://www.ids-imaging.us/store_us/u3-3990se-rev-1-2.html},
  note         = {Accessed: 2026-08-15}
}

@misc{thorlabsELL14,
  author       = {{Thorlabs, Inc.}},
  title        = {{ELL14 Elliptec Motorized Rotation Mount}},
  howpublished = {Product documentation},
  year         = {n.d.},
  url          = {https://www.thorlabs.com/newgrouppage9.cfm?objectgroup_id=12829},
  note         = {Accessed: 2026-08-15}
}

@book{adams2019optics,
  author    = {Adams, Charles S. and Hughes, Ifan G.},
  title     = {Optics f2f: From Fourier to Fresnel},
  publisher = {Oxford University Press},
  year      = {2019},
  doi       = {10.1093/oso/9780198786788.001.0001}
}

@inproceedings{oquab2023dinov2,
  title={DINOv2: Learning Robust Visual Features without Supervision},
  author={Oquab, Maxime and Darcet, Timoth{\'e}e and Moutakanni, Th{\'e}o and Vo, Huy and Szafraniec, Marc and Khalidov, Vasil and Fernandez, Pierre and Haziza, Daniel and Massa, Francisco and El-Nouby, Alaaeldin and Assran, Mahmoud and Ballas, Nicolas and Galuba, Wojciech and Howes, Russell and Huang, Po-Yao and Li, Shang-Wen and Misra, Ishan and Rabbat, Michael and Sharma, Vasu and Synnaeve, Gabriel and Xu, Hu and Jegou, Herv{\'e} and Mairal, Julien and Labatut, Patrick and Joulin, Armand and Bojanowski, Piotr},
  booktitle={Transactions on Machine Learning Research},
  year={2024}
}
}

\clearpage
\appendix
\renewcommand{\thetable}{S\arabic{table}}
\renewcommand{\theHtable}{supp.\arabic{table}}
\setcounter{table}{0}
\renewcommand{\thefigure}{S\arabic{figure}}
\renewcommand{\theHfigure}{supp.\arabic{figure}}
\setcounter{figure}{0}

\end{document}